\documentclass[10pt,twocolumn,letterpaper]{article}

 \usepackage{iccv}              

\usepackage{amsmath}
\usepackage{soul}
\usepackage[dvipsnames]{xcolor}
\usepackage{amssymb}
\usepackage{pifont}
\usepackage{adjustbox}
\usepackage{graphicx}
\usepackage{booktabs}
\usepackage{makecell}
\usepackage{mathtools}
\usepackage{appendix}
\usepackage[T1]{fontenc}

\newcommand{\x}{\ensuremath{\boldsymbol{x}}}

\newcommand{\vc}{\ensuremath{\boldsymbol{c}}}

\newcommand{\bbD}{\ensuremath{\mathbb{D}}}
\newcommand{\kl}{\ensuremath{\bbD_{\text{KL}}}}

\newcommand{\noise}{\ensuremath{\boldsymbol{\epsilon}}}

\definecolor{iccvblue}{rgb}{0.21,0.49,0.74}
\usepackage[pagebackref,breaklinks,colorlinks,allcolors=iccvblue]{hyperref}
\usepackage{multirow}
\usepackage{amssymb}

\def\confName{ICCV}
\def\confYear{2025}

\title{SUDO: Enhancing Text-to-Image Diffusion Models with \\ Self-Supervised Direct Preference Optimization}

\author{
Liang Peng$^{1}${\thanks{Equal contribution} \thanks{Work was done at FABU Inc.}} 
\quad Boxi Wu${^{2*}}$ 
\quad Haoran Cheng${^{2*}}$ 
\quad Yibo Zhao${^{1, 2}}$ 
\quad Xiaofei He${^{1, 2}}$  \\
\textsuperscript{\rm 1}FABU Inc.
\textsuperscript{\rm 2}Zhejiang University \\
{\tt\small  \{pengliang, wuboxi, chenghaoran\}@zju.edu.cn}
}

\begin{document}
\maketitle

\begin{abstract}
Previous text-to-image diffusion models typically employ supervised fine-tuning (SFT) to enhance pre-trained base models. 
However, this approach primarily minimizes the loss of mean squared error (MSE) at the pixel level, neglecting the need for global optimization at the image level, which is crucial for achieving high perceptual quality and structural coherence.
In this paper, we introduce \underline{S}elf-s\underline{U}pervised \underline{D}irect preference \underline{O}ptimization (SUDO), a novel paradigm that optimizes both fine-grained details at the pixel level and global image quality. 
By integrating direct preference optimization into the model, SUDO generates preference image pairs in a self-supervised manner, enabling the model to prioritize global-level learning  while complementing the pixel-level MSE loss.
As an effective alternative to supervised fine-tuning, SUDO can be seamlessly applied to any text-to-image diffusion model. Importantly, it eliminates the need for costly data collection and annotation efforts typically associated with traditional direct preference optimization methods. Through extensive experiments on widely-used models, including Stable Diffusion 1.5 and XL, we demonstrate that SUDO significantly enhances both global and local image quality.
The codes are provided at \href{https://github.com/SPengLiang/SUDO}{this link}.
\end{abstract}    
\section{Introduction}
\label{sec:intro}

The advancement of text-to-image diffusion models \cite{rombach2022high,podell2023sdxl,peebles2023scalable,esser2024scaling,xie2024sana,blackforestlabs2024flux}  has sparked significant progress in generative AI, particularly in producing high-quality images from textual descriptions. These models are typically trained on large datasets and can be fine-tuned on a high-quality dataset using supervised fine-tuning (SFT) \cite{dai2023emu,chen2023pixart}. 
SFT focuses on optimizing the model by minimizing the mean squared error (MSE) loss in the latent space, with the goal of refining the model's noise predictions at the per-pixel level. 
However, this fine-tuning process primarily addresses local discrepancies within the pixel, failing to account for the global optimization of the entire image. 
As a consequence, the model may struggle to generalize across diverse inputs, potentially leading to reduced image quality and poorer text-image alignment in generated outputs.

		\begin{figure}[t]
\centering 
		\includegraphics[trim=3 3 3 3, clip, width=1.0\linewidth]{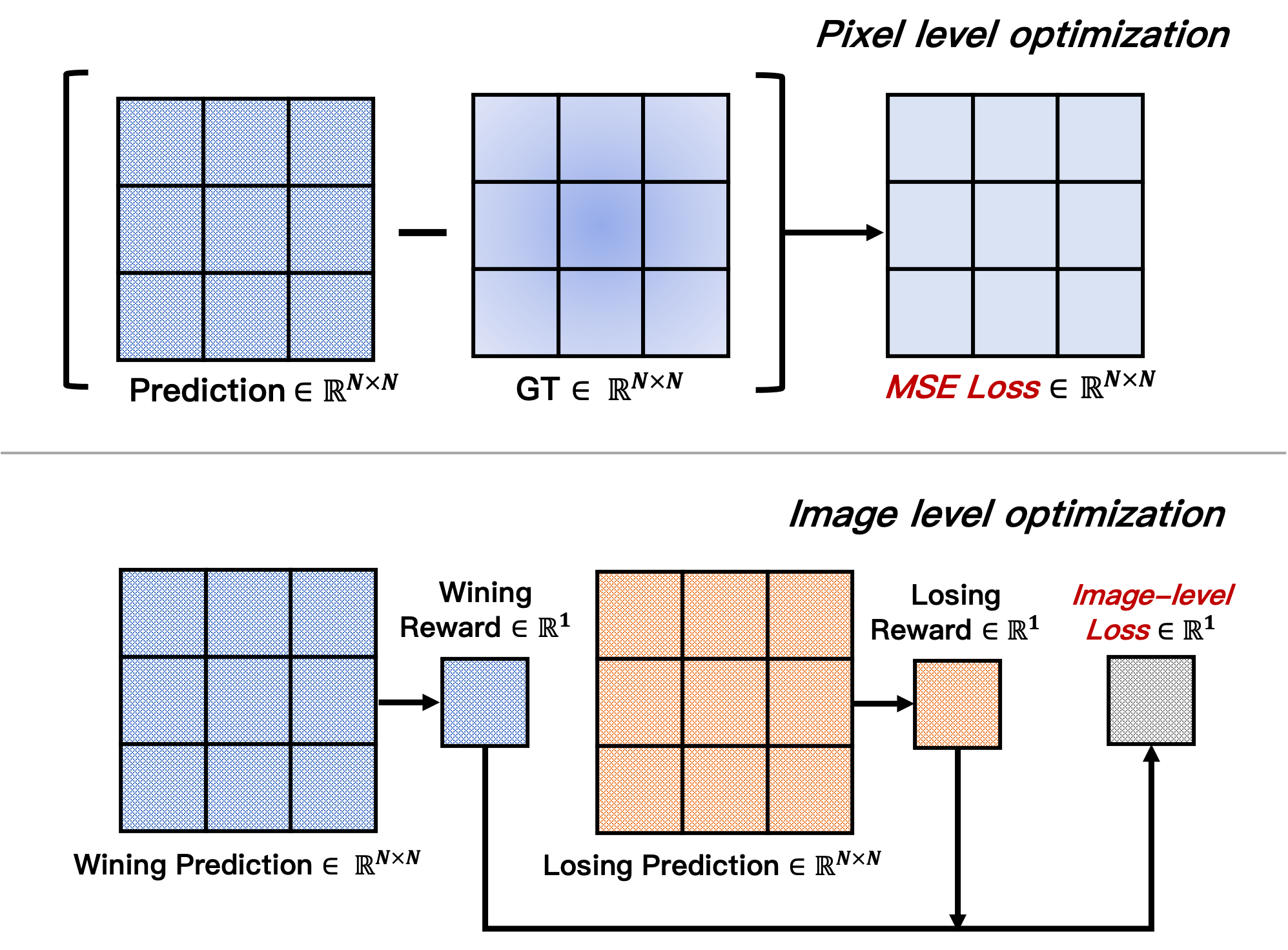}     
        \vspace{-7mm}
		\caption{
                    Pixel-level and image-level optimization.
                    In the conventional fine-tuning process, previous diffusion methods typically focus on pixel-level MSE loss. 
                    We enhance text-to-image diffusion models by incorporating image-level learning, achieved through self-supervised direct preference optimization.
				}
		\label{fig:level}
                \vspace{-4mm}
\end{figure}

		\begin{figure*}[t]
\centering 
		\includegraphics[trim=3 3 3 3, clip, width=1.0\linewidth]{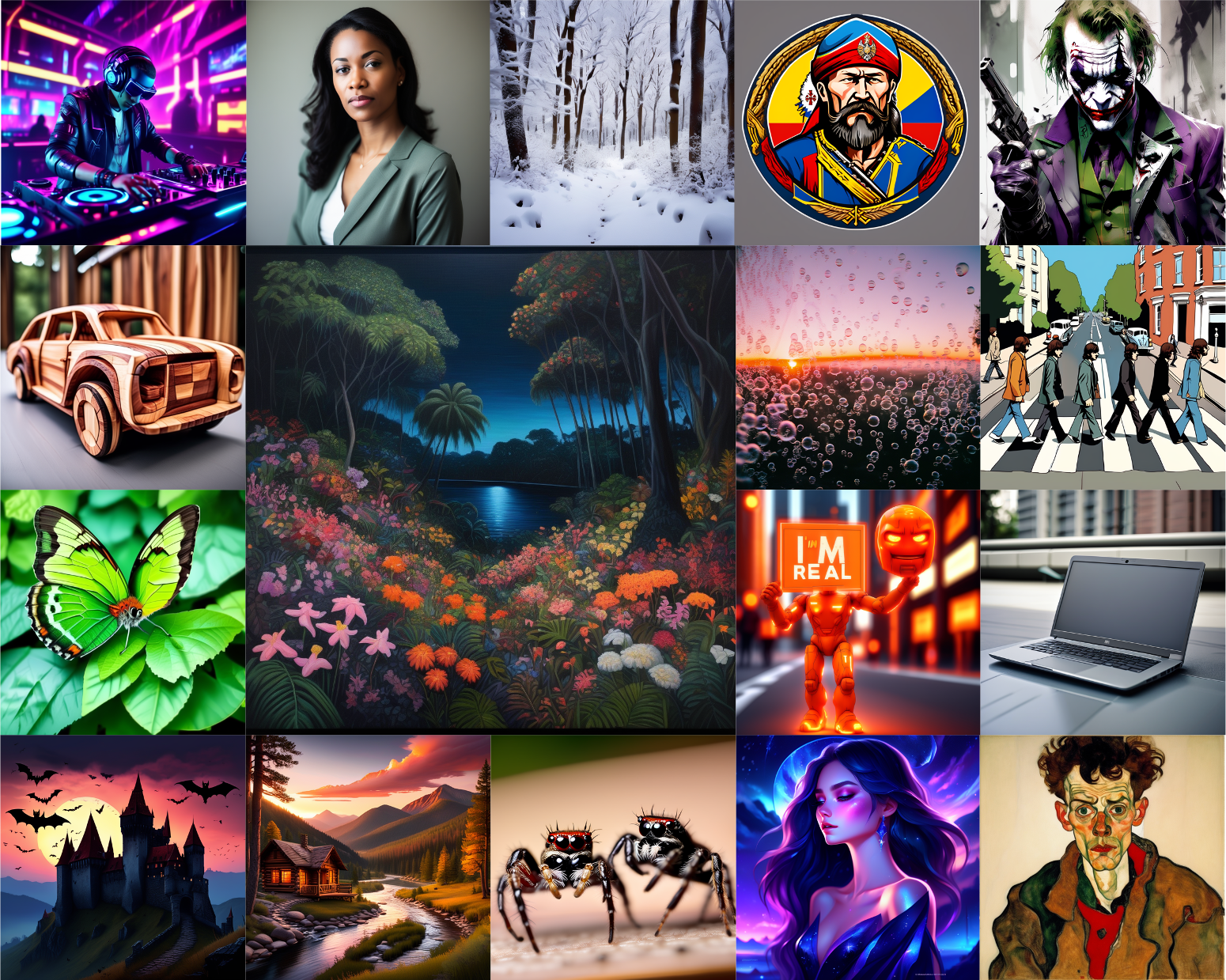}  
        \vspace{-7mm}
		\caption{
        We develop SUDO, a method for fine-tuning text-to-image diffusion models.
        It incorporates direct preference optimization in a self-supervised manner.
        We provide the results generated by the fine-tuned SDXL model with our method. 
	Best viewed in color.
				}
		\label{fig:head}
        \vspace{-3mm}
\end{figure*}

To mitigate these limitations, we propose \underline{\textbf{S}}elf-s\underline{\textbf{U}}pervised \underline{\textbf{D}}irect preference \underline{\textbf{O}}ptimization (SUDO), which incorporates global-level learning \cite{rafailov2023direct,wallace2024diffusion} in a self-supervised manner. In contrast to SFT’s focus on fine-grained pixel-level predictions through MSE loss, SUDO jointly optimizes the model at the global image level using preference image pairs. 
We illustrate the core idea of image-level learning in Figure \ref{fig:level}.
Such image pairs are generated by downgrading image labels in a self-supervised fashion, \textit{i.e.}, text-image perturbation. This strategy encourages the model to learn from both local and global contexts within the latent space, thereby capturing richer and more informative patterns. 
Unlike traditional direct preference optimization (DPO) methods \cite{rafailov2023direct,wallace2024diffusion}, which often rely on costly collections and human annotations, SUDO’s self-supervised design eliminates the need for expensive labeled datasets. 
We summarize the training data requirements in Table \ref{tab:data}.

\begin{table}[t]
\centering
\begin{tabular}{l|ccc}
				\toprule  
				Data requirments & SFT & SUDO & DPO \\ 
				\midrule    
				  Text Prompt & \checkmark & \checkmark &  \checkmark\\
                    Image & \checkmark & \checkmark &  \checkmark\\
				  Preference Image & -  & - &  \checkmark\\
					\bottomrule 
			\end{tabular}
                    \vspace{-3mm}
		\caption{ 
        Data requirements for different fine-tuning manners.
        The proposed SUDO shares the same data requirements as SFT.
        DPO needs extra preference images that are collected and annotated by human, which is highly time-consuming and expensive.
		}
                            \vspace{-4mm}
		\label{tab:data}
\end{table}

Importantly, SUDO is model-agnostic and can be applied to most text-to-image diffusion models. It offers an alternative to SFT, providing performance gains without extra manual annotations or dataset modifications. Our experiments on popular models such as Stable Diffusion 1.5 \cite{rombach2022high} and XL \cite{podell2023sdxl} show that SUDO yields substantial improvements in both overall image quality and text-image alignment. Furthermore, when compared to DPO approaches \cite{wallace2024diffusion}, SUDO demonstrates significant enhancements despite using only self-supervised preference images rather than manually annotated pairs, underscoring its potential to advance text-to-image diffusion models.
As shown in Figure \ref{fig:head}, we provide some visualization results using the fine-tuned SDXL model with SUDO.

In summary, we introduce SUDO as a self-supervised method to direct preference optimization, consistently enhancing model performance in text-to-image generation tasks. This manner opens promising avenues for future work and underscores the value of global-level learning.

\section{Related Work}
\label{sec:related_work}

\subsection{Text-to-Image Diffusion Models}
Text-to-image diffusion models have recently attracted significant attention in generative modeling due to their ability to produce high-quality and diverse images from textual descriptions. The introduction of Denoising Diffusion Probabilistic Models (DDPMs) \cite{dhariwal2021diffusion,ho2020denoising} marked a major breakthrough in this field, establishing diffusion models as a powerful generative approach. These models operate by reversing a gradual diffusion process, where noise is incrementally added to an image and then learned to be removed, ultimately synthesizing a coherent and realistic image. Building upon this foundation, diffusion models have become a representative paradigm in text-to-image generation, leading to models like GLIDE \cite{nichol2021glide} for text-guided image editing, Imagen \cite{saharia2022photorealistic} with cascaded diffusion for high-resolution synthesis, and Stable Diffusion \cite{rombach2022high} using latent diffusion for efficient generation.
To enhance image generation quality, researchers typically focus on two main directions. One approach involves architectural improvements, with the Diffusion Transformer (DiT) \cite{peebles2023scalable} gaining attention for its improved image fidelity, performance, and diversity. Notable advancements include PixArt-$\alpha$ \cite{chen2023pixart}, Hunyuan-DiT \cite{li2024hunyuan}, and SD3 \cite{esser2024scaling}.
Another approach leverages supervised fine-tuning to refine text-to-image diffusion models. These methods curate datasets by integrating various strategies, such as preference models \cite{podell2023sdxl}, pre-trained image models \cite{betker2023improving, dong2023raft, wu2023human} (e.g., image captioning models), and expert-assisted data filtering \cite{dai2023emu}.

\subsection{Preference-Based Optimization Methods}
In recent years, preference-based optimization has gained traction, refining models through user feedback or ranked preference pairs. 
In Large Language Models, Reinforcement Learning from Human Feedback (RLHF) leverages human comparisons to train a reward model that guides policy learning \cite{bai2022constitutional, ouyang2022training}. Alternatively, direct preference optimization (DPO) fine-tunes models directly on preference data, bypassing the need for an explicit reward model while achieving comparable performance \cite{rafailov2023direct}. 
Subsequently, preference-based optimization has been applied to image generation. Some methods enhance image quality by increasing rewards for preferred outputs \cite{clark2023directly,prabhudesai2023aligning,hao2023optimizing}, while others use reinforcement learning \cite{hao2023optimizing,black2023training}. However, training reliable reward models remains challenging and computationally expensive, with over-optimization potentially leading to mode collapse, reducing diversity \cite{lee2023aligning,prabhudesai2023aligning}.
Similarly, Direct Preference Optimization has been introduced in text-to-image generation, with Diffusion-DPO \cite{wallace2024diffusion} demonstrating the effectiveness of optimizing on human comparison data to enhance both visual appeal and text alignment. Additionally, direct score preference optimization \cite{zhudspo} refines diffusion models through score matching, providing a novel approach to preference learning. Several recent studies have further explored adapting preference learning techniques from large language models to fine-tune diffusion models \cite{yang2024using,li2025aligning,yuan2025self}, highlighting the growing interest in aligning generative models with human preferences.

\subsection{Self-Supervised Learning}
Self-supervised learning has emerged as a pivotal paradigm in machine learning. 
Among its most widely used approaches are contrastive self-supervised learning and generative self-supervised learning.
Contrastive self-supervised learning distinguishes representations by pulling similar instances closer while pushing dissimilar ones apart. MoCo \cite{he2020momentum} enhances training efficiency through a momentum encoder, while SimCLR \cite{chen2020simple} simplifies the process with strong data augmentations. CLIP \cite{radford2021learning} extends contrastive learning to vision-language tasks, aligning images and text in a shared latent space, enabling zero-shot transfer.
Generative approaches inherently follow unsupervised or self-supervised learning principles, training without labeled data to model the underlying distributions of the input. GANs \cite{goodfellow2014generative} utilize adversarial training to generate realistic data, while VAEs \cite{kingma2013auto} encode data into a structured latent space for controlled synthesis. VQ-VAE \cite{van2017neural} introduces discrete latent representations for high-quality generation. MAE \cite{he2022masked} leverages masked image modeling to learn rich visual features. More recently, denoising diffusion models \cite{ho2020denoising} have demonstrated impressive results by iteratively adding and removing noise, learning robust representations in a self-supervised manner.
Self-supervised learning enables training without manually labeled data, laying the foundation for future advancements in representation learning, generative modeling, and multimodal understanding.

\section{Methods}
\label{sec:methods}

\subsection{Preliminary}

{\bf{Diffusion Models.}}
Denoising Diffusion Probabilistic Models \cite{ho2020denoising} represent the image generation process as a Markovian process.
Let \( \x_0 \in \mathbb{R}^d \) be the data point and \( q(\x_t | \x_{t-1}) \) denote the forward process, where noise is added to the data at each timestep \( t \). The forward process is defined as:
\vspace{-3mm}
\begin{equation}
q(\x_t | \x_{t-1}) = \mathcal{N}(\x_t; \sqrt{1-\beta_t} \x_{t-1}, \beta_t \mathbb{I}),
\vspace{-2mm}
\end{equation}
where \( \beta_t \) is a schedule that controls the variance of noise added at each timestep, and \( \mathcal{N}(\cdot) \) denotes the normal distribution. The forward process gradually adds noise, with \( \x_T \) being pure noise after \( T \) timesteps.
The reverse process aims to learn the distribution \( p_\theta(\x_{t-1} | \x_t) \), which represents the process of denoising and generating data from pure noise. 
The reverse process can be parameterized as:
\vspace{-2mm}
\begin{equation}
p_\theta(\x_{t-1} | \x_t) = \mathcal{N}(\x_{t-1}; \mu_\theta(\x_t, t), \Sigma_\theta(\x_t, t)),
\vspace{-1mm}
\end{equation}
where \( \mu_\theta(\x_t, t) \) and \( \Sigma_\theta(\x_t, t) \) are the mean and covariance learned by the model at each timestep \( t \). The model is trained by minimizing the following objective:
\vspace{-1mm}
\begin{equation}
\mathcal{L}_{\text{diffusion}} = \mathbb{E}_{t, \x_0, \epsilon}\left[\| \epsilon - \epsilon_\theta(\x_t, t) \|^2\right],
\vspace{-1mm}
\end{equation}
where \( \epsilon \) is the noise added at each timestep, and \( \epsilon_\theta(\x_t, t) \) is the model's predicted noise. The model is trained to predict this noise accurately at each step, enabling it to reverse the diffusion process and generate high-quality data.
~\\
\noindent
{\bf{Reinforcement Learning from Human Feedback.}}
For diffusion models, human preferences at each diffusion step are modeled using a Bradley-Terry formulation \cite{bradley1952rank}, where the probability of preferring a “winning” sample \(\x_t^w\) over a “losing” sample \(\x_t^l\) for a given prompt \(\vc\) is defined as:
\begin{equation}
\vspace{-1mm}
p_\text{BT}(\x_t^w \succ \x_t^l | \vc) = \sigma\big(r(\vc, \x_t^w) - r(\vc, \x_t^l)\big),
\end{equation}
with \(\sigma, r, \vc\) representing the sigmoid function, reward model, and prompt, respectively. 
Subsequently, by conceptualizing the diffusion denoising process as a multi-step Markov Decision Process, the generative model is fine-tuned via reinforcement learning. 
The training objective \cite{wallace2024diffusion,fan2023reinforcement} is formulated as:
\begin{multline}
\vspace{-5mm}
 \mathcal{L}_{\text{rlhf}} = \mathbb{E}_{\vc\sim D}\mathbb{E}_{p_\theta(\x_{0:T}|\vc)} \sum\nolimits_{t=0}^{T-1} r(\vc, \x_t) - \\
 \lambda\,\kl\big(p_\theta(\x_{0:T}|\vc) \,\|\, p_\text{ref}(\x_{0:T}|\vc)\big),
\end{multline}
where $p_\text{ref}(\x_{0:T}|\vc)$ is the distribution from the pretrained diffusion model and \(\lambda\) controls the influence of the KL divergence regularization term. 
~\\
\noindent
{\bf{Direct Preference Optimization (DPO).}}
DPO streamlines RLHF by using the learning policy’s log likelihood to imp licitly encode the reward. In text-to-image diffusion models, this leads to a step-wise reward defined as:
\begin{equation}
r(\vc, \x_t) = \lambda \log \frac{p_\theta(\x_t \mid \x_{t+1}, \vc)}{p_{\text{ref}}(\x_t \mid \x_{t+1}, \vc)}.
\end{equation}
Recent works \cite{wallace2024diffusion,zhudspo} follow this line and adapt it to diffusion models. 
They optimize the model \(p_\theta\) based on the Bradley-Terry model \cite{bradley1952rank}, leading to an objective function:
\begin{multline}
 \mathcal{L}_{\text{Diffusion-DPO}} = -\mathbb{E}\Bigl[\log \sigma\Bigl(\lambda \log \frac{p_\theta(\x_t^{w} \mid \x_{t+1}^{w}, \vc)}
{p_{\text{ref}}(\x_t^{w} \mid \x_{t+1}^{w}, \vc)} \\
- \lambda \log \frac{p_\theta(\x_t^{l} \mid \x_{t+1}^{l}, \vc)}
{p_{\text{ref}}(\x_t^{l} \mid \x_{t+1}^{l}, \vc)}\Bigr) \Bigr],
\end{multline}
where the winning and losing samples \((\x_t^w, \x_t^l)\) and the prompt \(\vc\) are drawn from the dataset, and timestep \(t\) is uniformly sampled from the diffusion process. 
This formulation effectively aligns the learning policy with human preference signals embedded in the reference model.

\begin{figure}[t]
\centering 
		\includegraphics[trim=3 9 3 3, clip, width=1.0\linewidth]{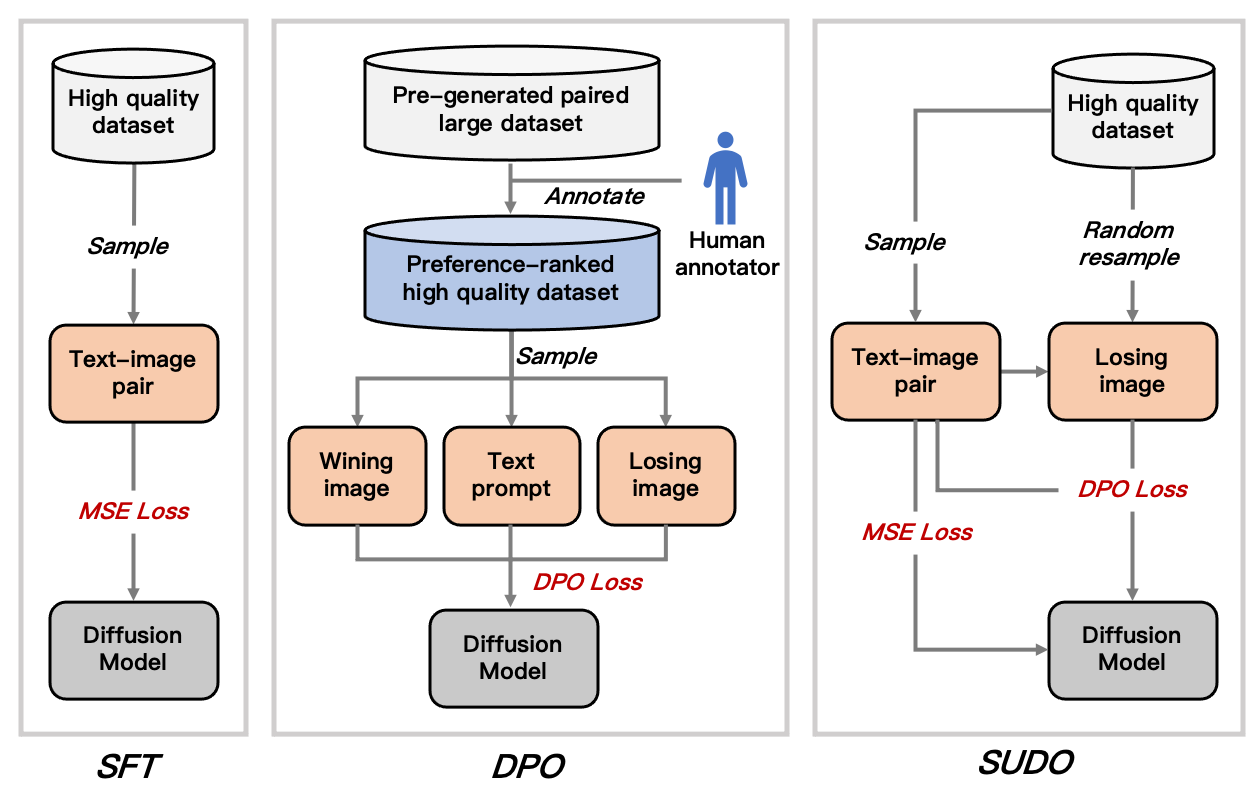} 
        \vspace{-7mm}
		\caption{
                    Different training process of SFT, DPO, and SUDO.
                    We generate losing images in a self-supervised manner, to perform direct  preference optimization under a regular dataset.
				Best viewed in color with zoom in.
				}
		\label{fig:method}
                \vspace{-3mm}
\end{figure}

\subsection{SUDO for Text-to-image Diffusion Models}
Inspired by DPO’s effective alignment with human preferences at the image level \cite{wallace2024diffusion}, our work aims to extend this formulation into the standard fine-tuning process for diffusion models. 
Unfortunately, a direct application of DPO is not feasible because it requires collecting image pairs (typically generated by different models with the same prompt or by using different seeds with the same model—along with their associated manual rankings). 
This approach does not align with the conventional fine-tuning pipeline and incurs significant additional costs. 
To overcome this limitation, we propose SUDO, which generates preference image pairs in a self-supervised manner.

\begin{figure*}[t]
\centering 
		\includegraphics[trim=3 3 3 3, clip, width=1.0\linewidth]{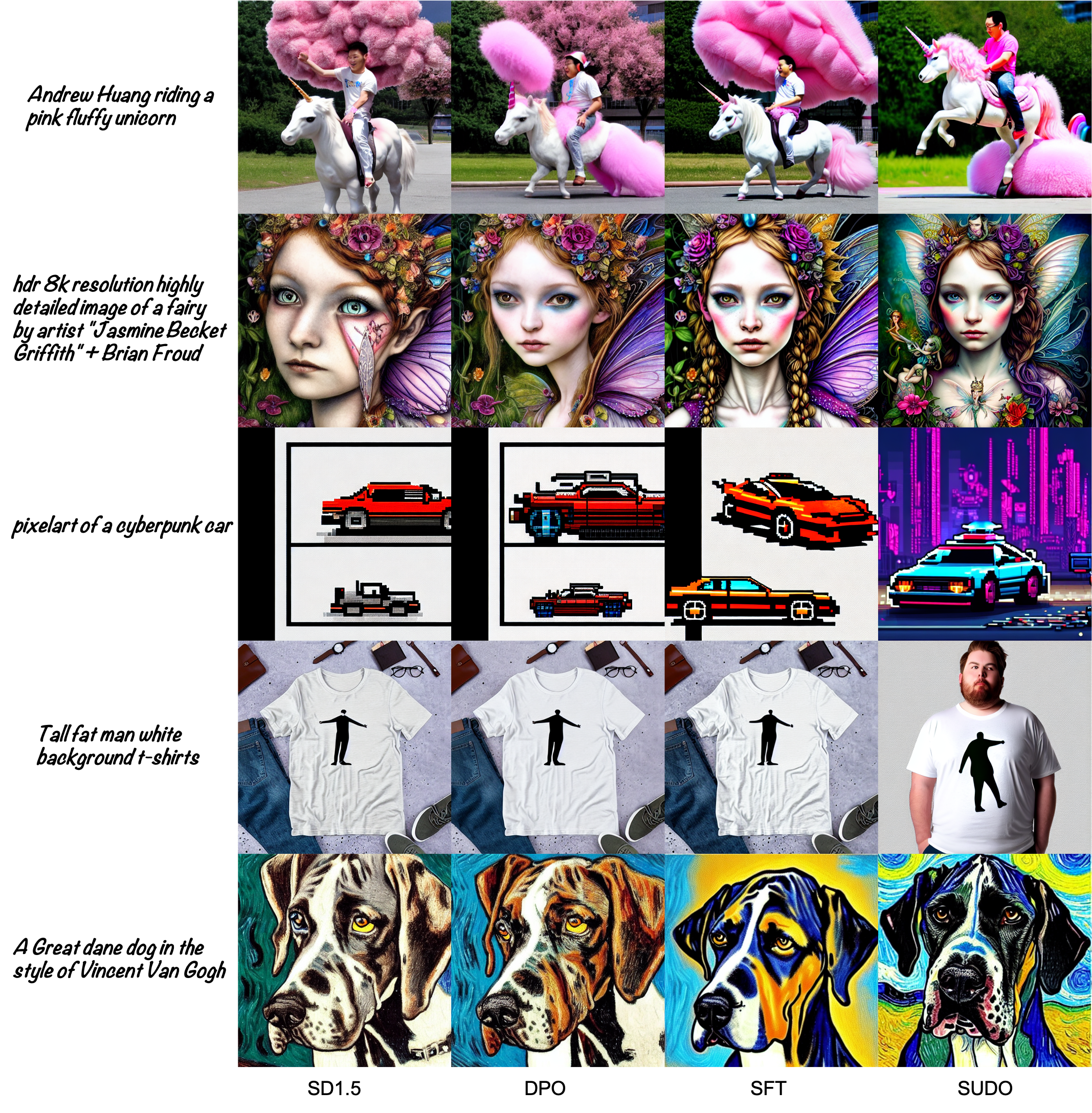}   
            \vspace{-9mm}
		\caption{
         Qualitative comparisons with the SD1.5 base model. 
         All results are generated with the same random seed.
         Comparing with SFT and DPO \cite{wallace2024diffusion}, the model trained by SUDO demonstrates superior text prompt alignment.
         It also shows more appealing visual quality, especially in layout, colors, and details.
	   Best viewed in color.
				}
		\label{fig:sd15}
                    \vspace{-2mm}
\end{figure*}

In each training iteration, we denote the text-image pair as \((\vc, \x)\), where the image component is regarded as the wining image \(\x^w\). 
Traditional DPO methods \cite{wallace2024diffusion,rafailov2023direct} require generating image pairs corresponding to the same prompt, followed by manual selection of the preferred (wining) and less preferred (losing) images, denoted as \(\x^w\) and \(\x^l\), respectively. 
In contrast, our method eliminates the need for a manually selected the losing images. Instead, we obtain "self-generated losing samples" by systematically degrading the winner \(\x^w\). 
Formally, the losing image is represented as \(\x^{sl} = \text{\textbf{Downgrade}}(\x^w)\).
Inspired by \cite{wallace2024diffusion}, the self-supervised direct preference loss is:
\vspace{-2mm}
\begin{multline}
    \mathcal{L}_{\text{SUDO}} = - \log \sigma \Bigg( 
     C \Big[ \big\| \noise_\theta(\x_{t}^w,t) - \noise^w \big\|_2^2 - \big\| \noise_\theta(\x_{t}^{sl},t) - \noise^{sl} \big\|_2^2 \\
    - \big( \big\| \noise_\text{ref}(\x_{t}^w,t) - \noise^w \big\|_2^2 - \big\| \noise_\text{ref}(\x_{t}^{sl},t) - \noise^{sl} \big\|_2^2 \big) \Big] 
    \Bigg), \\
    \text{where} \quad \x^{sl} = \text{\textbf{Downgrade}}(\x^w)
    \label{eq:loss-sudo}
\vspace{-1mm}
\end{multline}
where $C$ and $\noise^{sl}$ refer to a scale factor and the noise corresponding to losing images, and $\noise_\text{ref}$ is the reference model.
In our implementation, the \(\text{\textbf{Downgrade}}\) operation is simply performed by randomly selecting images from the training dataset. 
Namely, for each self-generated image pair, the winning sample closely aligns with the prompt, whereas the losing sample fails to correspond to the description.
We encourage future work to explore more degradation strategies for self-supervised direct preference optimization. 
The training process overview is shown in Figure \ref{fig:method}.
s

\textbf{Overall Loss.} Concerning the above discussion, the final loss is as follows:
\vspace{-2mm}
\begin{equation}
\mathcal{L}= \lambda_1\mathcal{L}_{\text{MSE}} + \lambda_2\mathcal{L}_{\text{SUDO}}
\vspace{-1mm}
\end{equation}
We empirically set $\lambda_1$ and $\lambda_2$ to $0.5, 0.5$, respectively.
$\mathcal{L}_{\text{MSE}}$ makes the model focus on pixel level optimization, and $\mathcal{L}_{\text{SUDO}}$ encougrages the model to learn information on the image level.
Such two losses are complementary.
\section{Experiments}
\label{sec:exp}

\begin{figure*}[t]
\centering 
		\includegraphics[width=1.0\linewidth]{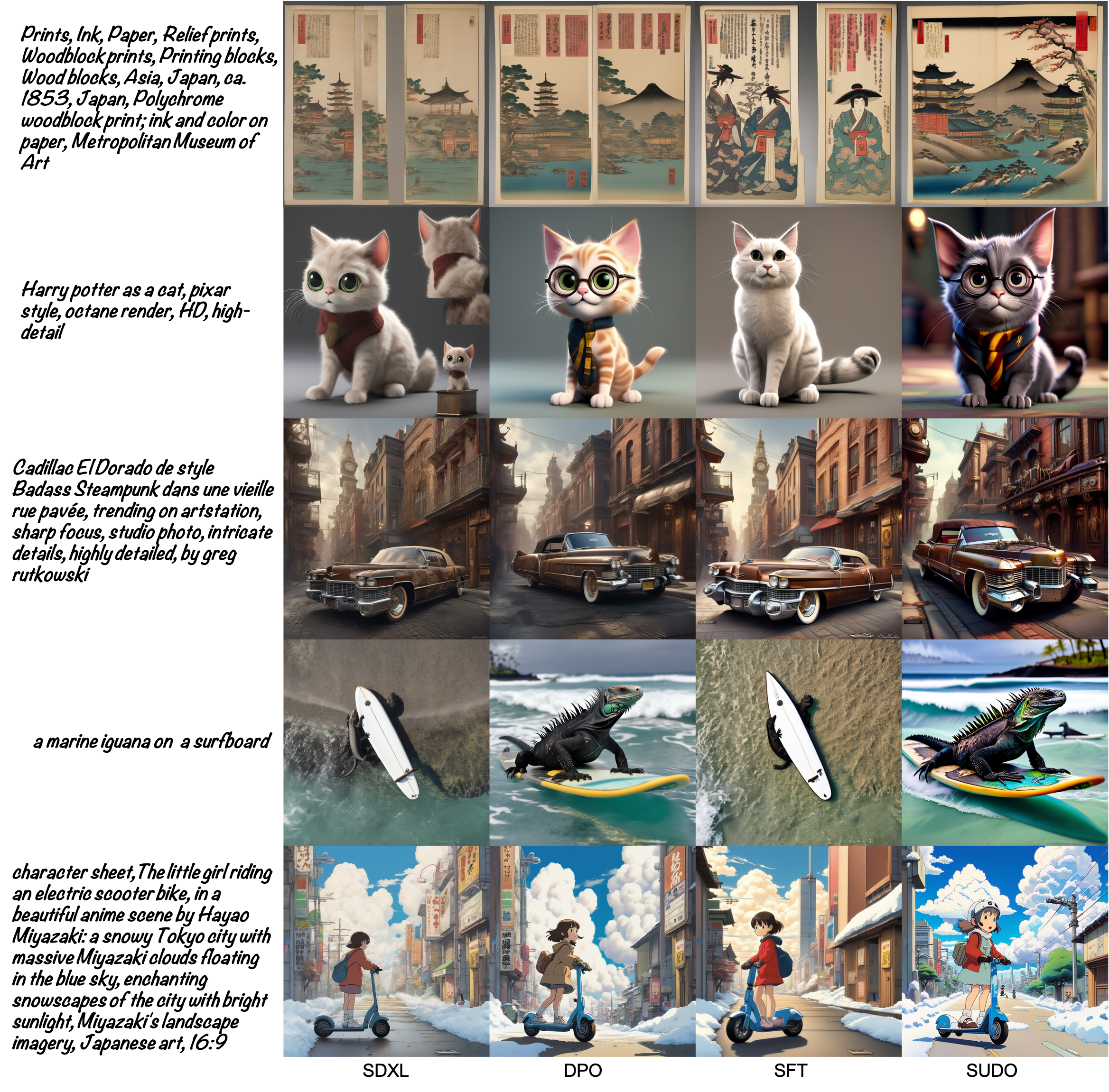}      
        \vspace{-8mm}
		\caption{
         Qualitative comparisons with the SDXL base model. 
        All results are generated using the same random seed. 
        Please note that the training dataset (Pick-a-Pic V2 \cite{kirstain2023pick}) used for fine-tuning is obtained from a SD1.5 variant, SD2.1, and SDXL models. 
        Consequently, directly fine-tuning (SFT) on this dataset does not lead to improvements—and may even result in degraded performance (\textit{e.g.}, as shown in the second row). 
        In contrast, DPO \cite{wallace2024diffusion} optimizes the model by leveraging preference relationships with pre-ranked pairs, thereby avoiding this issue. 
        SUDO uses the same data requirements as SFT but yields significant improvements in both text prompt alignment and visual quality, demonstraing its effectiveness.
        Best viewed in color.
       }
		\label{fig:sdxl}
\vspace{-1mm}
\end{figure*}

\subsection{Setup}

\begin{table*}[t]
\centering
\resizebox{\textwidth}{!}{%
\begin{tabular}{c|c|c|ccccc|ccccc}
				\toprule  
				\multirow{2}{*}{Datasets}&\multicolumn{2}{c|}{\multirow{2}{*}{Methods}}& \multicolumn{5}{c|}{SD1.5} & \multicolumn{5}{c}{SDXL} \\ 
                \cmidrule(lr){4-13}
                \multicolumn{1}{c|}{}&\multicolumn{2}{c|}{}& P.S. &  Aes.&  CLIP&  HPS&  I.R. & P.S. &  Aes. &  CLIP&  HPS&  I.R. \\ 
				\midrule    
				     \multirow{7}{*}{\shortstack{Pick-a-Pic \\ V2}} & Base& \multirow{4}{*}{\shortstack{Avg\\~\\score}}& 20.57  & 53.15 & 32.58 & 26.17 & -14.81 & 22.10 & \bf 60.01 & 35.86 & 26.83 & 50.62 \\
				    ~  &  SFT& ~  &21.10  & \bf 56.35 & 33.75 &27.03 & 45.03 &21.48 & 57.84 & 35.67 & 26.67 & 30.89  \\
                      ~  & DPO  & ~ &20.91  & 54.07 & 33.19 & 26.46 & 4.13 & \bf22.57 & 59.93 & 37.30 & 27.30 & 81.14 \\
                     ~  &  SUDO& ~  & \bf  21.23  &\bf  56.35 & \bf  34.79 & \bf  27.33 & \bf  71.00 & 22.34  & 59.97 &\bf  37.53 &\bf  27.89 & \bf  103.96 \\
                      \cmidrule(lr){2-13}
                      ~  & SFT& \multirow{3}{*}{\shortstack{Win\\~\\rate}} &75.00  &  77.20 & 60.40 & 90.20 & 80.00  &19.40  & 31.80 & 47.00 & 44.60 & 42.4 \\
                     ~  &  DPO& ~  &73.80  & 60.00 & 60.00  & 71.80 & 61.00 & \bf  72.60  & 47.20 & \bf  63.00  & 79.80 & 69.8 \\
                     ~  &  SUDO& ~  &\bf  78.60  & \bf 77.80 & \bf  68.40&\bf   94.20  & \bf  85.20 & 60.80  &\bf  50.80 & 62.40 &\bf  93.80 & \bf 79.2  \\
					\bottomrule 
                      \multirow{7}{*}{\shortstack{PartiPrompts}} & Base& \multirow{4}{*}{\shortstack{Avg\\~\\score}} & 21.39  & 53.13 & 33.21 & 26.79 & 1.48 &   22.63 & 57.69 & 35.77 & 27.33 & 69.78 \\
				      ~  &SFT & ~ &21.75  & \bf  55.31 & 33.93 & 27.57 & 50.75& 22.02  & 56.41 & 35.31 & 27.13 & 47.29\\
                       ~  &DPO  & ~&21.61  & 53.58 & 33.88 & 26.98  &21.43 & \bf  22.90 &  57.85 & 36.95 & 27.73 & 103.36\\
                       ~  &SUDO & ~ & \bf  21.84  & 55.09 & \bf  35.11 & \bf  27.84 &\bf  75.66& 22.79  & \bf  58.69 &\bf 37.00 &\bf  28.30 & \bf  117.50 \\
                       \cmidrule(lr){2-13}
                       ~  &SFT& \multirow{3}{*}{\shortstack{Win\\~\\rate}} &67.28  & \bf  70.89 & 53.43 & 85.42 & 73.35 &21.38  & 38.11 & 45.10 & 43.75 & 40.93\\
                      ~  & DPO & ~ &67.10 & 57.17 & 56.74 & 61.83 & 63.05 & \bf  63.42 & 53.62 & \bf  62.32 & 73.10 & 68.44\\
                      ~  & SUDO& ~  & \bf  69.85  & 68.50 & \bf  63.24 & \bf  89.40  & \bf 81.00 & 56.19  &\bf 60.48 & 60.17 &\bf 92.16  & \bf  76.84\\
                      \bottomrule
                     \multirow{7}{*}{\shortstack{HPD V2}} & Base & \multirow{4}{*}{\shortstack{Avg\\~\\score}} &20.84 & 54.32 & 33.96 & 26.84 & -11.79 &  22.78 & 61.34 & 37.68 & 27.68 &78.27 \\
				     ~  & SFT & ~  & 21.57  &\bf  57.41 & 35.26 & 27.89 & 57.74 & 22.24  & 60.08 & 37.39 & 27.76 & 66.62\\
                       ~  &DPO & ~  &21.30  &55.80 & 34.68 & 27.22 & 13.24 & \bf  23.18 & \bf 61.35 & \bf  38.45 & 28.14 & 102.74\\
                      ~  & SUDO & ~  &\bf  21.58 & 57.10 & \bf  36.30 & \bf  28.11 &\bf  76.13  & 22.98  & 61.30 &  38.35 &\bf  28.77 & \bf  110.67 \\
                       \cmidrule(lr){2-13}
                       ~  &SFT& \multirow{3}{*}{\shortstack{Win\\~\\rate}} &\bf  79.53 &\bf  75.31 & 59.34  & 90.10 & 81.16 &23.47  & 37.28 & 46.63 & 58.22 & 45.81 \\
                      ~  & DPO & ~  &75.72  & 66.28 & 57.56 & 72.43 & 64.69 & \bf  72.66 & \bf  50.28 & \bf  58.69 & 80.56 & 69.78\\
                      ~  & SUDO  & ~ &\bf 79.53  & 74.03 &\bf  68.47 &\bf  92.49 & \bf  85.19& 58.78  & 48.06 & 55.65 &\bf   94.97 & \bf   72.50 \\
					\bottomrule 
			\end{tabular}}
            \vspace{-2mm}
		\caption{ 
		Quantitative comparisons. 
            We compare different fine-tuning methods (SFT, DPO \cite{wallace2024diffusion}, and SUDO) on SD 1.5 and SDXL base models over three datasets (Pick-a-Pic V2 \cite{kirstain2023pick}, PartiPrompts \cite{yu2022scaling}, and HPDv2 \cite{wu2023human}).
            "P.S." refers to PickScore, "Aes." is Aesthetics, and "I.R." denotes ImageAward.s
            For the SD1.5 base model, our method achieves the best performance across most metrics. 
            In contrast, for the SDXL base model, we observe that SFT clearly degrades performance. 
            This is likely due to the fact that the training dataset (Pick-a-pic V2 \cite{kirstain2023pick}) used for fine-tuning is derived from SD1.5 variant, SD2.1, and SDXL models. 
            Interestingly, our method still achieves competitive results compared to DPO, which utilizes the full dataset and optimizes the model by leveraging preference relationships with pre-ranked pairs. 
            These results underscore the effectiveness and robustness of our approach.
		}
		\label{tab:main}
\end{table*}	
\noindent
{\bf{Implement Details:}}
Following Diffusion-DPO \cite{wallace2024diffusion}, for the SD1.5 \cite{rombach2022high} experiments, AdamW \cite{loshchilov2017decoupled} is utilized, while SDXL \cite{podell2023sdxl} training is conducted with Adafactor \cite{shazeer2018adafactor} to conserve memory.
Following the official implementation in Diffusion-DPO \cite{wallace2024diffusion}, $C$ in Equation \ref{eq:loss-sudo} is set to $-2500$.
For SD 1.5, a batch size of 2048 pairs (resolution: $512*512$) is maintained by training across 4 NVIDIA A100 GPU.
Each handling 8 pairs locally with gradient accumulation over 64 steps. 
For SDXL, considering the resource limitation, we use the total batch size of 96 pairs (resolution: $1024*1024$). 
Training is performed at fixed square resolutions. 
We use a learning rate of 1e-6 coupled with a 25\% linear warmup. 

~\\
\noindent
{\bf{Training Dataset:}}
Our training data is sourced from the Pick-a-Pic V2 dataset\cite{kirstain2023pick}, which keeps the same to Diffusion-DPO \cite{wallace2024diffusion}. 
It contains pairwise preference annotations for images generated by Dreamlike (a fine-tuned variant of SD1.5), SD2.1, and SDXL.
These prompts and preferences were collected from users of the Pick-a-Pic web application. 
\textbf{\textit{Please note that SUDO only uses the preference image and associated text in the dataset, instead of using the whole manually annotated image pairs.}}

~\\
\noindent
{\bf{Evaluation:}}
We conduct evaluation on three datasets: Pick-a-Pic V2 \cite{kirstain2023pick} validation set (contains 500 prompts), PartiPrompts \cite{yu2022scaling} (contains 1632 prompts, including diverse categories and challenge aspects), and HPDv2 \cite{wu2023human} (contains 3200 prompts, including anime, concept art, paintings and photo).
We compare SUDO with three different type baselines, \textit{i.e.}, base models (Stable Diffusion 1.5 (SD1.5) and Stable Diffusion XL (SDXL)), SFT models, DPO models, where DPO models are the publicly released from Diffusion-DPO \cite{wallace2024diffusion}.
For evaluation metrics, we employ the popular PickScore \cite{kirstain2023pick} , Aesthetics \cite{schuhmann2022laion}, CLIP \cite{radford2021learning}, HPS V2 \cite{wu2023human} , and ImageAward \cite{xu2023imagereward} scores.
For convenience of comparison, we scale the scores to fit a similar range (\textit{$\times$100 for PickScore, CLIP, HPS, and ImageAward, $\times$10 for Aesthetics}). 
\textbf{PickScore}
is a caption-sensitive scoring model, originally trained on Pick-a-Pic (v1), that estimates the perceived image quality by humans. 
\textbf{Aesthetics} 
assesses the visual appeal of an image, considering factors such as lighting, color harmony, composition, and overall artistic quality.
\textbf{CLIP} 
measures the semantic alignment between an image and a corresponding text prompt. By computing the cosine similarity between the image and text embeddings, this score evaluates how well the image content matches the provided textual description. 
\textbf{Human Preference Score (HPS V2)} 
is a metric designed to align with human judgments of image quality, particularly in the context of text-to-image synthesis.
\textbf{ImageAward} 
quantifies the quality, aesthetic appeal, or alignment of a generated image with respect to desired attributes. It is typically derived from a reward model trained on human preference data.

\subsection{Quantitative Results}s
    We provide quantitative comparisons in Table \ref{tab:main}.
    We compare our method with SFT and DPO \cite{wallace2024diffusion}. For the SD1.5 base model, our method significantly outperforms the alternatives. For example, SUDO achieves a CLIP score of 34.79 on the Pick-a-Pic V2 dataset—an improvement of +2.21 over the base model—whereas SFT and DPO yield improvements of +1.17 and +0.61, respectively. In terms of overall human preference metrics, SUDO delivers substantially higher gains, improving the base model from –14.81 to 71.00 on the ImageReward metric, which far exceeds the improvements observed with SFT (45.03) and DPO (4.13). Notably, the win rate of SUDO reaches 94.20 on the HPS metric. Results on other datasets (PartiPrompts and HPD v2) further confirm these improvements.
    In contrast, results on the SDXL base model show a slightly different scenario, particularly with SFT. We observe that SFT substantially degrades the performance of the base model. This degradation is likely due to the fact that, while the training dataset (comprising generations from an SD1.5 variant, SD2.1, and SDXL) is of much higher quality than that used for SD1.5, it does not exhibit clear superiority for SDXL. Nevertheless, the model trained by SUDO still demonstrates significantly better performance compared to both SFT and the base model. On the PickScore, Aesthetics, and CLIP metrics, SUDO achieves results comparable to DPO, and it attains superior scores on the HPS and ImageReward metrics. These findings validate our hypothesis that image-level learning benefits text-to-image diffusion models and highlight the superiority of SUDO.

\subsection{Qualitative Results}
    We provide qualitative comparisons in Figure \ref{fig:sd15} and Figure \ref{fig:sdxl} for SD1.5 and SDXL, respectively. All quantitative and qualitative results are generated using the same random seed. We observe that SUDO consistently improves over other models, particularly in terms of producing more vivid colors and better adherence to text prompts. For example, in the third row of Figure \ref{fig:sd15}, only SUDO successfully reveals the concept "cyberpunk". In the fourth row of Figure \ref{fig:sdxl}, both DPO and SUDO capture the intended meaning of the prompt, but SUDO exhibits a more appealing visual quality. These quantitative and qualitative results confirm the effectiveness of our approach.

\subsection{Ablation}
    We conduct ablation studies on various downgrade approaches, as shown in Table \ref{tab:down}. In particular, we examine two additional methods—namely, blur (which involves downsampling and upsampling the image by a factor of 4) and random grid (which divides the image into an $8\times8$ grid and randomly swaps two grids). We observe that these two downgrade methods lead to worse performance than SFT, indicating that significantly degraded image quality is harmful for image-level preference learning. 
    Additionally, we remove the MSE loss and report the results in Table \ref{tab:mse}. Interestingly, the performance does not drop noticeably, further confirming the effectiveness of image-level learning and the SUDO approach.

\begin{table}[t]
\centering
\resizebox{\linewidth}{!}{%
\begin{tabular}{l|ccccc}
				\toprule  
				Methods& P.S. &  Aes. &  CLIP&  HPS&  I.R. \\ 
				\midrule    
				    Base model& 20.57  & 53.15 & 32.58 & 26.17 & -14.81 \\
                      SFT &21.10  & \bf 56.35 & 33.75 &27.03 & 45.03 \\
				    w/ Blur  & 20.80  & 55.16 & 33.22 & 26.57 & 4.26 \\
                    w/ Random grid  & 20.85  & 55.86 & 32.71 & 26.64 & 24.21\\
                    w/ Random image & \bf 21.23  &\bf 56.35 & \bf 34.79 & \bf 27.33 & \bf 71.00 \\
					\bottomrule 
			\end{tabular}}
            \vspace{-3mm}
			        		\caption{ 
		Ablation on different downgrade manners.s
            Only the "Random image" row achieves significant improvements over SFT, implying that degrading the image quality is detrimental.
		}
		\label{tab:down}
        \vspace{-1mm}
\end{table}

\begin{table}[t]
\centering
\resizebox{\linewidth}{!}{%
\begin{tabular}{l|ccccc}
				\toprule  
				Methods& P.S.&  Aes.&  CLIP&  HPS&  I.R. \\ 
				\midrule    
				    Base model& 20.57  & 53.15 & 32.58 & 26.17 & -14.81 \\
                    SFT &21.10  & \bf 56.35 & 33.75 &27.03 & 45.03 \\
                   SUDO w/o MSE &  \bf 21.26  & 56.22 & \bf 35.00 & 27.31 & 67.78 \\
                    SUDO  &21.23  &\bf 56.35 &  34.79 & \bf 27.33 & \bf 71.00\\
					\bottomrule 
			\end{tabular}}
            \vspace{-3mm}
			        		\caption{ 
				Ablation on MSE loss.
            When removing MSE loss, SUDO still shows the superiority to SFT. 
		}
                    \vspace{-2mm}
		\label{tab:mse}
\end{table}

\section{Limitation and Future Work}
Our image-level learning is conducted using direct preference learning, which incurs additional computational cost during training—specifically, two forward passes (one for the winning image and one for the losing image) are required in each iteration.
Furthermore, we currently downgrade the winning image by randomly selecting images from the training dataset. However, when the training dataset is very small, this strategy may prove ineffective. 
We encourage future work to explore more effective self-supervised methods for generating losing images. 

\section{Conclusion}s
In this paper, we propose incorporating image-level learning into the fine-tuning process for text-to-image diffusion tasks, which emphasizes global optimization. 
To address this challenge, we introduce SUDO, a self-supervised approach based on direct preference optimization. 
In each training iteration, SUDO generates preference image pairs in a self-supervised manner and subsequently performs image-level learning.
We conduct experiments on various datasets using different base models, demonstrating the effectiveness of the proposed method.

{
    \small
    \bibliographystyle{ieeenat_fullname}
    \bibliography{main}

\begin{thebibliography}{44}
\providecommand{\natexlab}[1]{#1}
\providecommand{\url}[1]{\texttt{#1}}
\expandafter\ifx\csname urlstyle\endcsname\relax
  \providecommand{\doi}[1]{doi: #1}\else
  \providecommand{\doi}{doi: \begingroup \urlstyle{rm}\Url}\fi

\bibitem[Bai et~al.(2022)Bai, Kadavath, Kundu, Askell, Kernion, Jones, Chen,
  Goldie, Mirhoseini, McKinnon, et~al.]{bai2022constitutional}
Yuntao Bai, Saurav Kadavath, Sandipan Kundu, Amanda Askell, Jackson Kernion,
  Andy Jones, Anna Chen, Anna Goldie, Azalia Mirhoseini, Cameron McKinnon,
  et~al.
\newblock Constitutional ai: Harmlessness from ai feedback.
\newblock \emph{arXiv preprint arXiv:2212.08073}, 2022.

\bibitem[Betker et~al.(2023)Betker, Goh, Jing, Brooks, Wang, Li, Ouyang,
  Zhuang, Lee, Guo, et~al.]{betker2023improving}
James Betker, Gabriel Goh, Li Jing, Tim Brooks, Jianfeng Wang, Linjie Li, Long
  Ouyang, Juntang Zhuang, Joyce Lee, Yufei Guo, et~al.
\newblock Improving image generation with better captions.
\newblock \emph{Computer Science. https://cdn. openai. com/papers/dall-e-3.
  pdf}, 2\penalty0 (3):\penalty0 8, 2023.

\bibitem[Black et~al.(2023)Black, Janner, Du, Kostrikov, and
  Levine]{black2023training}
Kevin Black, Michael Janner, Yilun Du, Ilya Kostrikov, and Sergey Levine.
\newblock Training diffusion models with reinforcement learning.
\newblock \emph{arXiv preprint arXiv:2305.13301}, 2023.

\bibitem[Bradley and Terry(1952)]{bradley1952rank}
Ralph~Allan Bradley and Milton~E Terry.
\newblock Rank analysis of incomplete block designs: I. the method of paired
  comparisons.
\newblock \emph{Biometrika}, 39\penalty0 (3/4):\penalty0 324--345, 1952.

\bibitem[Chen et~al.(2023)Chen, Yu, Ge, Yao, Xie, Wu, Wang, Kwok, Luo, Lu,
  et~al.]{chen2023pixart}
Junsong Chen, Jincheng Yu, Chongjian Ge, Lewei Yao, Enze Xie, Yue Wu, Zhongdao
  Wang, James Kwok, Ping Luo, Huchuan Lu, et~al.
\newblock Pixart-$alpha$: Fast training of diffusion transformer for
  photorealistic text-to-image synthesis.
\newblock \emph{arXiv preprint arXiv:2310.00426}, 2023.

\bibitem[Chen et~al.(2020)Chen, Kornblith, Norouzi, and Hinton]{chen2020simple}
Ting Chen, Simon Kornblith, Mohammad Norouzi, and Geoffrey Hinton.
\newblock A simple framework for contrastive learning of visual
  representations.
\newblock In \emph{International conference on machine learning}, pages
  1597--1607. PmLR, 2020.

\bibitem[Clark et~al.(2023)Clark, Vicol, Swersky, and Fleet]{clark2023directly}
Kevin Clark, Paul Vicol, Kevin Swersky, and David~J Fleet.
\newblock Directly fine-tuning diffusion models on differentiable rewards.
\newblock \emph{arXiv preprint arXiv:2309.17400}, 2023.

\bibitem[Dai et~al.(2023)Dai, Hou, Ma, Tsai, Wang, Wang, Zhang, Vandenhende,
  Wang, Dubey, et~al.]{dai2023emu}
Xiaoliang Dai, Ji Hou, Chih-Yao Ma, Sam Tsai, Jialiang Wang, Rui Wang, Peizhao
  Zhang, Simon Vandenhende, Xiaofang Wang, Abhimanyu Dubey, et~al.
\newblock Emu: Enhancing image generation models using photogenic needles in a
  haystack.
\newblock \emph{arXiv preprint arXiv:2309.15807}, 2023.

\bibitem[Dhariwal and Nichol(2021)]{dhariwal2021diffusion}
Prafulla Dhariwal and Alexander Nichol.
\newblock Diffusion models beat gans on image synthesis.
\newblock \emph{Advances in neural information processing systems},
  34:\penalty0 8780--8794, 2021.

\bibitem[Dong et~al.(2023)Dong, Xiong, Goyal, Zhang, Chow, Pan, Diao, Zhang,
  Shum, and Zhang]{dong2023raft}
Hanze Dong, Wei Xiong, Deepanshu Goyal, Yihan Zhang, Winnie Chow, Rui Pan,
  Shizhe Diao, Jipeng Zhang, Kashun Shum, and Tong Zhang.
\newblock Raft: Reward ranked finetuning for generative foundation model
  alignment.
\newblock \emph{arXiv preprint arXiv:2304.06767}, 2023.

\bibitem[Esser et~al.(2024)Esser, Kulal, Blattmann, Entezari, M{\"u}ller,
  Saini, Levi, Lorenz, Sauer, Boesel, et~al.]{esser2024scaling}
Patrick Esser, Sumith Kulal, Andreas Blattmann, Rahim Entezari, Jonas
  M{\"u}ller, Harry Saini, Yam Levi, Dominik Lorenz, Axel Sauer, Frederic
  Boesel, et~al.
\newblock Scaling rectified flow transformers for high-resolution image
  synthesis.
\newblock In \emph{Forty-first international conference on machine learning},
  2024.

\bibitem[Fan et~al.(2023)Fan, Watkins, Du, Liu, Ryu, Boutilier, Abbeel,
  Ghavamzadeh, Lee, and Lee]{fan2023reinforcement}
Ying Fan, Olivia Watkins, Yuqing Du, Hao Liu, Moonkyung Ryu, Craig Boutilier,
  Pieter Abbeel, Mohammad Ghavamzadeh, Kangwook Lee, and Kimin Lee.
\newblock Reinforcement learning for fine-tuning text-to-image diffusion
  models.
\newblock In \emph{Thirty-seventh Conference on Neural Information Processing
  Systems (NeurIPS) 2023}. Neural Information Processing Systems Foundation,
  2023.

\bibitem[Goodfellow et~al.(2014)Goodfellow, Pouget-Abadie, Mirza, Xu,
  Warde-Farley, Ozair, Courville, and Bengio]{goodfellow2014generative}
Ian Goodfellow, Jean Pouget-Abadie, Mehdi Mirza, Bing Xu, David Warde-Farley,
  Sherjil Ozair, Aaron Courville, and Yoshua Bengio.
\newblock Generative adversarial nets.
\newblock \emph{Advances in neural information processing systems}, 27, 2014.

\bibitem[Hao et~al.(2023)Hao, Chi, Dong, and Wei]{hao2023optimizing}
Yaru Hao, Zewen Chi, Li Dong, and Furu Wei.
\newblock Optimizing prompts for text-to-image generation.
\newblock \emph{Advances in Neural Information Processing Systems},
  36:\penalty0 66923--66939, 2023.

\bibitem[He et~al.(2020)He, Fan, Wu, Xie, and Girshick]{he2020momentum}
Kaiming He, Haoqi Fan, Yuxin Wu, Saining Xie, and Ross Girshick.
\newblock Momentum contrast for unsupervised visual representation learning.
\newblock In \emph{Proceedings of the IEEE/CVF conference on computer vision
  and pattern recognition}, pages 9729--9738, 2020.

\bibitem[He et~al.(2022)He, Chen, Xie, Li, Doll{\'a}r, and
  Girshick]{he2022masked}
Kaiming He, Xinlei Chen, Saining Xie, Yanghao Li, Piotr Doll{\'a}r, and Ross
  Girshick.
\newblock Masked autoencoders are scalable vision learners.
\newblock In \emph{Proceedings of the IEEE/CVF conference on computer vision
  and pattern recognition}, pages 16000--16009, 2022.

\bibitem[Ho et~al.(2020)Ho, Jain, and Abbeel]{ho2020denoising}
Jonathan Ho, Ajay Jain, and Pieter Abbeel.
\newblock Denoising diffusion probabilistic models.
\newblock \emph{Advances in neural information processing systems},
  33:\penalty0 6840--6851, 2020.

\bibitem[Kingma et~al.(2013)Kingma, Welling, et~al.]{kingma2013auto}
Diederik~P Kingma, Max Welling, et~al.
\newblock Auto-encoding variational bayes, 2013.

\bibitem[Kirstain et~al.(2023)Kirstain, Polyak, Singer, Matiana, Penna, and
  Levy]{kirstain2023pick}
Yuval Kirstain, Adam Polyak, Uriel Singer, Shahbuland Matiana, Joe Penna, and
  Omer Levy.
\newblock Pick-a-pic: An open dataset of user preferences for text-to-image
  generation.
\newblock \emph{Advances in Neural Information Processing Systems},
  36:\penalty0 36652--36663, 2023.

\bibitem[Labs(2024)]{blackforestlabs2024flux}
Black~Forest Labs.
\newblock Flux, 2024.
\newblock Black Forest Labs. Flux, 2024.

\bibitem[Lee et~al.(2023)Lee, Liu, Ryu, Watkins, Du, Boutilier, Abbeel,
  Ghavamzadeh, and Gu]{lee2023aligning}
Kimin Lee, Hao Liu, Moonkyung Ryu, Olivia Watkins, Yuqing Du, Craig Boutilier,
  Pieter Abbeel, Mohammad Ghavamzadeh, and Shixiang~Shane Gu.
\newblock Aligning text-to-image models using human feedback.
\newblock \emph{arXiv preprint arXiv:2302.12192}, 2023.

\bibitem[Li et~al.(2025)Li, Kallidromitis, Gokul, Kato, and
  Kozuka]{li2025aligning}
Shufan Li, Konstantinos Kallidromitis, Akash Gokul, Yusuke Kato, and Kazuki
  Kozuka.
\newblock Aligning diffusion models by optimizing human utility.
\newblock \emph{Advances in Neural Information Processing Systems},
  37:\penalty0 24897--24925, 2025.

\bibitem[Li et~al.(2024)Li, Zhang, Lin, Xiong, Long, Deng, Zhang, Liu, Huang,
  Xiao, et~al.]{li2024hunyuan}
Zhimin Li, Jianwei Zhang, Qin Lin, Jiangfeng Xiong, Yanxin Long, Xinchi Deng,
  Yingfang Zhang, Xingchao Liu, Minbin Huang, Zedong Xiao, et~al.
\newblock Hunyuan-dit: A powerful multi-resolution diffusion transformer with
  fine-grained chinese understanding.
\newblock \emph{arXiv preprint arXiv:2405.08748}, 2024.

\bibitem[Loshchilov and Hutter(2017)]{loshchilov2017decoupled}
Ilya Loshchilov and Frank Hutter.
\newblock Decoupled weight decay regularization.
\newblock \emph{arXiv preprint arXiv:1711.05101}, 2017.

\bibitem[Nichol et~al.(2021)Nichol, Dhariwal, Ramesh, Shyam, Mishkin, McGrew,
  Sutskever, and Chen]{nichol2021glide}
Alex Nichol, Prafulla Dhariwal, Aditya Ramesh, Pranav Shyam, Pamela Mishkin,
  Bob McGrew, Ilya Sutskever, and Mark Chen.
\newblock Glide: Towards photorealistic image generation and editing with
  text-guided diffusion models.
\newblock \emph{arXiv preprint arXiv:2112.10741}, 2021.

\bibitem[Ouyang et~al.(2022)Ouyang, Wu, Jiang, Almeida, Wainwright, Mishkin,
  Zhang, Agarwal, Slama, Ray, et~al.]{ouyang2022training}
Long Ouyang, Jeffrey Wu, Xu Jiang, Diogo Almeida, Carroll Wainwright, Pamela
  Mishkin, Chong Zhang, Sandhini Agarwal, Katarina Slama, Alex Ray, et~al.
\newblock Training language models to follow instructions with human feedback.
\newblock \emph{Advances in neural information processing systems},
  35:\penalty0 27730--27744, 2022.

\bibitem[Peebles and Xie(2023)]{peebles2023scalable}
William Peebles and Saining Xie.
\newblock Scalable diffusion models with transformers.
\newblock In \emph{Proceedings of the IEEE/CVF international conference on
  computer vision}, pages 4195--4205, 2023.

\bibitem[Podell et~al.(2023)Podell, English, Lacey, Blattmann, Dockhorn,
  M{\"u}ller, Penna, and Rombach]{podell2023sdxl}
Dustin Podell, Zion English, Kyle Lacey, Andreas Blattmann, Tim Dockhorn, Jonas
  M{\"u}ller, Joe Penna, and Robin Rombach.
\newblock Sdxl: Improving latent diffusion models for high-resolution image
  synthesis.
\newblock \emph{arXiv preprint arXiv:2307.01952}, 2023.

\bibitem[Prabhudesai et~al.(2023)Prabhudesai, Goyal, Pathak, and
  Fragkiadaki]{prabhudesai2023aligning}
Mihir Prabhudesai, Anirudh Goyal, Deepak Pathak, and Katerina Fragkiadaki.
\newblock Aligning text-to-image diffusion models with reward backpropagation.
\newblock 2023.

\bibitem[Radford et~al.(2021)Radford, Kim, Hallacy, Ramesh, Goh, Agarwal,
  Sastry, Askell, Mishkin, Clark, et~al.]{radford2021learning}
Alec Radford, Jong~Wook Kim, Chris Hallacy, Aditya Ramesh, Gabriel Goh,
  Sandhini Agarwal, Girish Sastry, Amanda Askell, Pamela Mishkin, Jack Clark,
  et~al.
\newblock Learning transferable visual models from natural language
  supervision.
\newblock In \emph{International conference on machine learning}, pages
  8748--8763. PmLR, 2021.

\bibitem[Rafailov et~al.(2023)Rafailov, Sharma, Mitchell, Manning, Ermon, and
  Finn]{rafailov2023direct}
Rafael Rafailov, Archit Sharma, Eric Mitchell, Christopher~D Manning, Stefano
  Ermon, and Chelsea Finn.
\newblock Direct preference optimization: Your language model is secretly a
  reward model.
\newblock \emph{Advances in Neural Information Processing Systems},
  36:\penalty0 53728--53741, 2023.

\bibitem[Rombach et~al.(2022)Rombach, Blattmann, Lorenz, Esser, and
  Ommer]{rombach2022high}
Robin Rombach, Andreas Blattmann, Dominik Lorenz, Patrick Esser, and Bj{\"o}rn
  Ommer.
\newblock High-resolution image synthesis with latent diffusion models.
\newblock In \emph{Proceedings of the IEEE/CVF conference on computer vision
  and pattern recognition}, pages 10684--10695, 2022.

\bibitem[Saharia et~al.(2022)Saharia, Chan, Saxena, Li, Whang, Denton,
  Ghasemipour, Gontijo~Lopes, Karagol~Ayan, Salimans,
  et~al.]{saharia2022photorealistic}
Chitwan Saharia, William Chan, Saurabh Saxena, Lala Li, Jay Whang, Emily~L
  Denton, Kamyar Ghasemipour, Raphael Gontijo~Lopes, Burcu Karagol~Ayan, Tim
  Salimans, et~al.
\newblock Photorealistic text-to-image diffusion models with deep language
  understanding.
\newblock \emph{Advances in Neural Information Processing Systems},
  35:\penalty0 36479--36494, 2022.

\bibitem[Schuhmann(2022)]{schuhmann2022laion}
Christoph Schuhmann.
\newblock Laion-aesthetics.
\newblock \url{https://laion.ai/blog/laion-aesthetics/}, 2022.
\newblock Accessed: 2023-11-10.

\bibitem[Shazeer and Stern(2018)]{shazeer2018adafactor}
Noam Shazeer and Mitchell Stern.
\newblock Adafactor: Adaptive learning rates with sublinear memory cost.
\newblock In \emph{International Conference on Machine Learning}, pages
  4596--4604. PMLR, 2018.

\bibitem[Van Den~Oord et~al.(2017)Van Den~Oord, Vinyals, et~al.]{van2017neural}
Aaron Van Den~Oord, Oriol Vinyals, et~al.
\newblock Neural discrete representation learning.
\newblock \emph{Advances in neural information processing systems}, 30, 2017.

\bibitem[Wallace et~al.(2024)Wallace, Dang, Rafailov, Zhou, Lou, Purushwalkam,
  Ermon, Xiong, Joty, and Naik]{wallace2024diffusion}
Bram Wallace, Meihua Dang, Rafael Rafailov, Linqi Zhou, Aaron Lou, Senthil
  Purushwalkam, Stefano Ermon, Caiming Xiong, Shafiq Joty, and Nikhil Naik.
\newblock Diffusion model alignment using direct preference optimization.
\newblock In \emph{Proceedings of the IEEE/CVF Conference on Computer Vision
  and Pattern Recognition}, pages 8228--8238, 2024.

\bibitem[Wu et~al.(2023)Wu, Hao, Sun, Chen, Zhu, Zhao, and Li]{wu2023human}
Xiaoshi Wu, Yiming Hao, Keqiang Sun, Yixiong Chen, Feng Zhu, Rui Zhao, and
  Hongsheng Li.
\newblock Human preference score v2: A solid benchmark for evaluating human
  preferences of text-to-image synthesis.
\newblock \emph{arXiv preprint arXiv:2306.09341}, 2023.

\bibitem[Xie et~al.(2024)Xie, Chen, Chen, Cai, Tang, Lin, Zhang, Li, Zhu, Lu,
  et~al.]{xie2024sana}
Enze Xie, Junsong Chen, Junyu Chen, Han Cai, Haotian Tang, Yujun Lin, Zhekai
  Zhang, Muyang Li, Ligeng Zhu, Yao Lu, et~al.
\newblock Sana: Efficient high-resolution image synthesis with linear diffusion
  transformers.
\newblock \emph{arXiv preprint arXiv:2410.10629}, 2024.

\bibitem[Xu et~al.(2023)Xu, Liu, Wu, Tong, Li, Ding, Tang, and
  Dong]{xu2023imagereward}
Jiazheng Xu, Xiao Liu, Yuchen Wu, Yuxuan Tong, Qinkai Li, Ming Ding, Jie Tang,
  and Yuxiao Dong.
\newblock Imagereward: Learning and evaluating human preferences for
  text-to-image generation.
\newblock \emph{Advances in Neural Information Processing Systems},
  36:\penalty0 15903--15935, 2023.

\bibitem[Yang et~al.(2024)Yang, Tao, Lyu, Ge, Chen, Shen, Zhu, and
  Li]{yang2024using}
Kai Yang, Jian Tao, Jiafei Lyu, Chunjiang Ge, Jiaxin Chen, Weihan Shen,
  Xiaolong Zhu, and Xiu Li.
\newblock Using human feedback to fine-tune diffusion models without any reward
  model.
\newblock In \emph{Proceedings of the IEEE/CVF Conference on Computer Vision
  and Pattern Recognition}, pages 8941--8951, 2024.

\bibitem[Yu et~al.(2022)Yu, Xu, Koh, Luong, Baid, Wang, Vasudevan, Ku, Yang,
  Ayan, et~al.]{yu2022scaling}
Jiahui Yu, Yuanzhong Xu, Jing~Yu Koh, Thang Luong, Gunjan Baid, Zirui Wang,
  Vijay Vasudevan, Alexander Ku, Yinfei Yang, Burcu~Karagol Ayan, et~al.
\newblock Scaling autoregressive models for content-rich text-to-image
  generation.
\newblock \emph{arXiv preprint arXiv:2206.10789}, 2\penalty0 (3):\penalty0 5,
  2022.

\bibitem[Yuan et~al.(2025)Yuan, Chen, Ji, and Gu]{yuan2025self}
Huizhuo Yuan, Zixiang Chen, Kaixuan Ji, and Quanquan Gu.
\newblock Self-play fine-tuning of diffusion models for text-to-image
  generation.
\newblock \emph{Advances in Neural Information Processing Systems},
  37:\penalty0 73366--73398, 2025.

\bibitem[Zhu et~al.()Zhu, Xiao, and Honavar]{zhudspo}
Huaisheng Zhu, Teng Xiao, and Vasant~G Honavar.
\newblock Dspo: Direct score preference optimization for diffusion model
  alignment.
\newblock In \emph{The Thirteenth International Conference on Learning
  Representations}.

\end{thebibliography}
}

\clearpage
\appendix
\renewcommand{\thesection}{}

\onecolumn

\begin{center}
\textbf{\Large Appendix for SUDO: Enhancing Text-to-Image Diffusion Models with \\ Self-Supervised Direct Preference Optimization}
\end{center}
\setcounter{figure}{0}
\setcounter{page}{1}

\begin{figure}[h]
\centering 
	\includegraphics[width=1.0\linewidth]{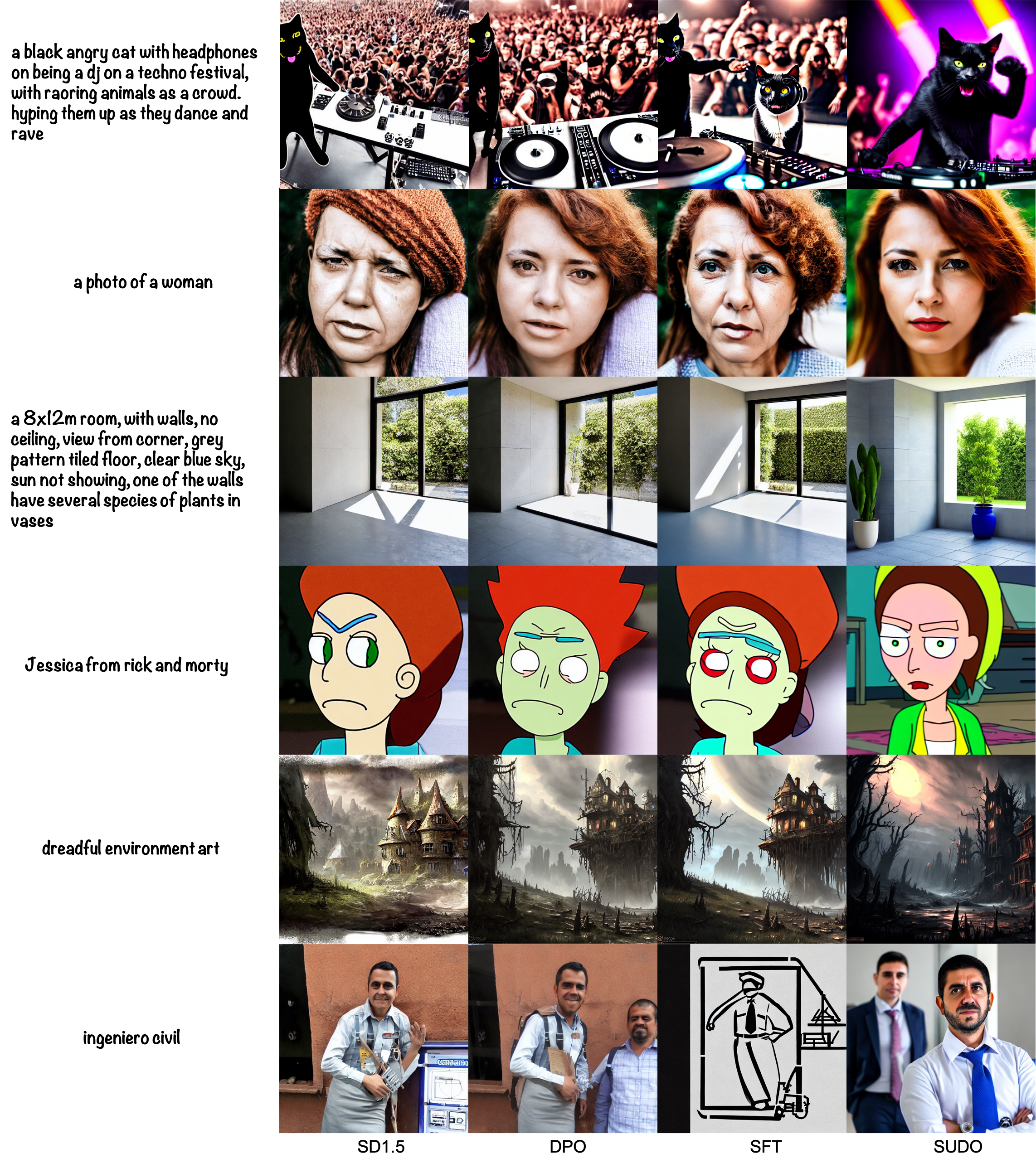}      
	\caption{Extra qualitative comparisons with the SD1.5 base model.}
	\label{fig:sdxl}
\end{figure}

\begin{figure*}[t]
\centering 
		\includegraphics[width=1.0\linewidth]{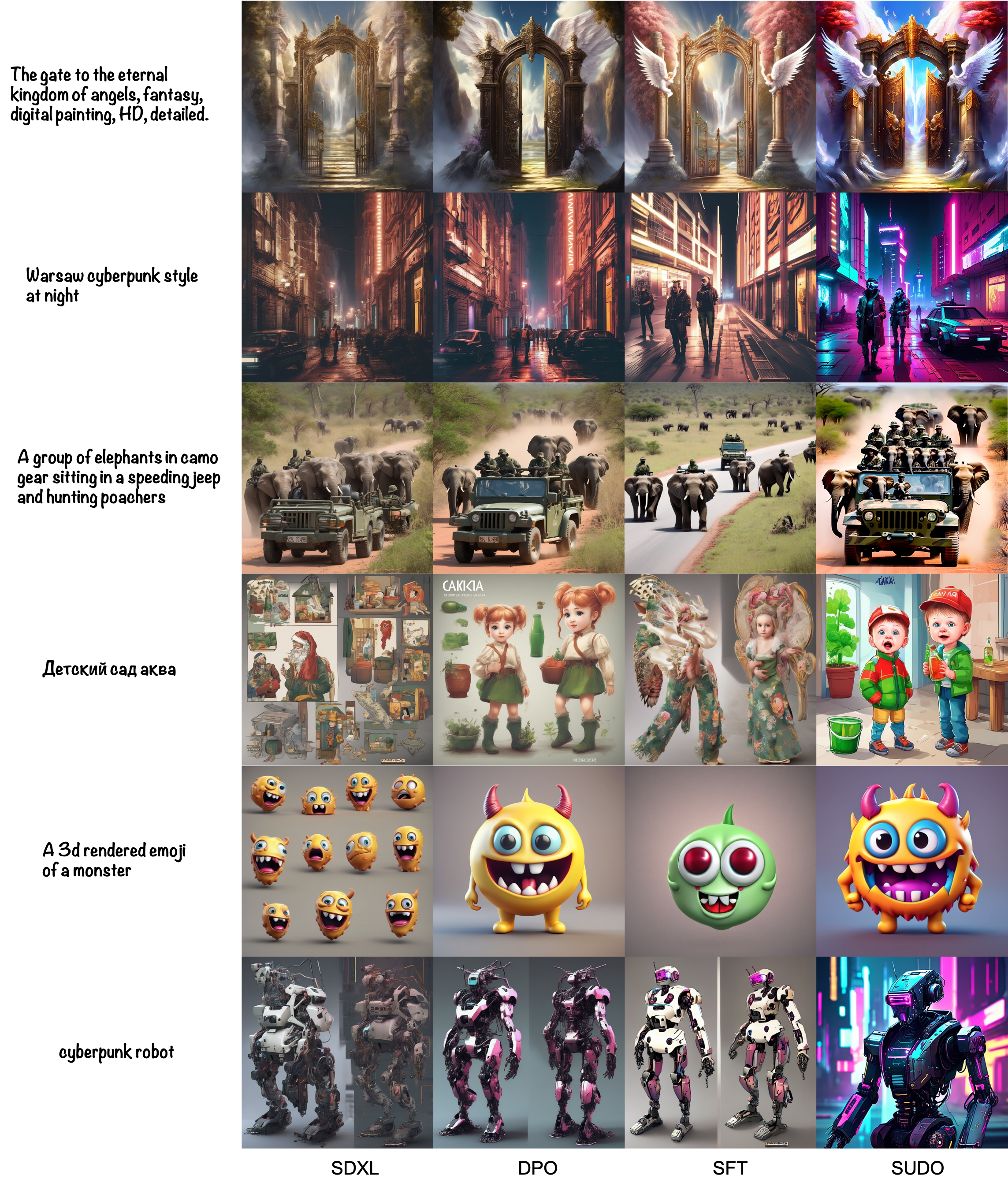}      
		\caption{
         Extra qualitative comparisons with the SDXL base model. }
\end{figure*}

\end{document}